\documentclass[]{aastex701}
\usepackage{amsmath}
\usepackage{txfonts}
\usepackage{bm}
\usepackage[ruled]{algorithm2e}
\usepackage{color}
\usepackage{comment}

\newcommand{\x}{t}
\newcommand{\y}{{\bf y}}
\newcommand{\f}{{\bf f}}
\newcommand*{\params}{\ensuremath{\boldsymbol{\theta}}\xspace}

\begin{document}

\title{Data-driven informative priors for Bayesian inference with quasi-periodic data}

\author[orcid=0000-0003-2402-8166,sname='Lopez-Santiago']{Javier L\'opez-Santiago}
\altaffiliation{}
\affiliation{Universidad Carlos III de Madrid, Department of Signal Theory and Communications, Avenida de la Universidad 30 (edificio Sabatini), 28911, Legan\'es (Madrid), Spain}
\email[show]{jalopezs@ing.uc3m.es}  

\author[orcid=0000-0002-7611-6558,sname='Martino']{Luca Martino} 
\altaffiliation{}
\affiliation{Universit\`a degli Studi di Catania, Corso Italia 55, Catania, Italy}
\email[]{lukatotal@gmail.com}

\author[orcid=0000-0001-5227-7253,sname='Miguez']{Joaqu\'in Miguez}
\altaffiliation{}
\affiliation{Universidad Carlos III de Madrid, Department of Signal Theory and Communications, Avenida de la Universidad 30 (edificio Sabatini), 28911, Legan\'es (Madrid), Spain}
\email[]{jmiguez@ing.uc3m.es}

\author[orcid=0000-0002-5793-1557,sname='Vazquez-Vilar']{Gonzalo V\'azquez-Vilar}
\altaffiliation{}
\affiliation{Universidad Carlos III de Madrid, Department of Signal Theory and Communications, Avenida de la Universidad 30 (edificio Sabatini), 28911, Legan\'es (Madrid), Spain}
\email[]{govazquez@ing.uc3m.es}

\begin{abstract}

{Bayesian computational strategies for inference can be inefficient in approximating the posterior distribution in models that exhibit some form of periodicity.} This is because the probability mass of the marginal posterior distribution of the parameter representing the period is usually highly concentrated in a very small region of the parameter space. Therefore, it is necessary to provide as much information as possible to the inference method through the parameter prior distribution. We intend to show that it is possible to construct {a prior distribution} from the data by fitting a Gaussian process (GP) with a periodic kernel. More specifically, we want to show that it is possible to approximate the marginal posterior distribution of the hyperparameter corresponding to the period in the kernel. Subsequently, this distribution can be used as a prior distribution for the inference method. We use an adaptive importance sampling method to approximate the posterior distribution of the hyperparameters of the GP. Then, we use the marginal posterior distribution of the hyperparameter related to the periodicity in order to construct a prior distribution for the period of the parametric model. 
{This workflow is empirical Bayes, implemented as a modular (\emph{cut}) transfer of a GP posterior for the period to the parametric model.}
We {applied} the proposed methodology to both synthetic and real data. {We approximated the posterior distribution of the period of the GP kernel and then passed it forward as a posterior-as-prior with no feedback. Finally, we analyzed} its impact on the marginal posterior distribution.  


\end{abstract}



\section{Introduction} 

Bayesian {statistics} provides a robust framework for incorporating prior knowledge into probabilistic analyses of underlying processes. When combined with Monte Carlo methods, it facilitates the approximation of probability distributions for random variables (for example, model parameters) conditioned on observed data and/or specified models. More specifically, Bayesian methods are able to unveil the uncertainty in the parameters of interest by providing prior information and iteratively improving estimates based on observed data.
{Numerical inference} algorithms work well in a reduced parameter space. However, they become less efficient as the space to be covered increases. The latter occurs when the dimension of the parameter space (i.e. the number of parameters to infer) is large, or when the posterior probability distribution is highly concentrated. This is a common situation in problems of inference from periodic or quasi-periodic models, where the probability mass of the period is highly concentrated. Therefore, it is necessary to reduce the length of the parameter space to ensure the posterior probability distribution is correctly sampled. This is achieved by restricting the width of the prior distribution, i.e. by constructing more informative 
priors. {Note that here we use the term \emph{informative prior} in its most generic sense to refer to a prior that reduces the extension of the parameter space by incorporating more knowledge, in contrast to non-informative or diffuse priors. In some specialized Bayesian literature, informative priors are considered subjective \citep[e.g.][]{press2009}. Prior distributions incorporate specific knowledge or beliefs about parameters. Consequently, their specification is inherently subjective. However, incorporating this information can improve the accuracy and efficiency of the inference \citep{bernardo1985}.}

The use of informative priors has been long discussed in many fields \citep{clarke1996, lenk2009, morita2010, pedroza2018, finch2019} and there is a consensus about the convenience of choosing informative priors instead of diffuse priors when possible. For example, in the literature we can find several contributions where the authors have proposed the use of data-driven or model-based approaches to define priors for model selection \citep[see the survey][for more details]{SafePriors}. In general, data-driven prior construction can be useful for {computational} Bayesian inference\footnote{The term "Bayesian inference" is used generically here, even though it is combined with data-driven prior construction.} \citep{Martin2019,holden2022,ravasi2024}. 
This strategy reduces the need for subjective and arbitrary prior specification but there are other important advantages. From a computational point of view, the construction of prior densities that are not in conflict with the main modes of the likelihood function helps Monte Carlo algorithms to achieve good performance. That is, this approach can improve the accuracy and robustness of statistical inference. In contrast, the use of diffuse priors in model selection tends to reduce the importance of the analyzed model \citep[provided that the area under the likelihood is finite;][]{SafePriors,Martino25fake}. More specifically, the marginal likelihood of the model tends to zero as the prior becomes less informative, making model selection more difficult. The model would be automatically discarded in the selection by imposing arbitrary diffuse priors (such as uniform pdfs with a large variance). Hence, for model selection problems, a diffuse prior becomes very informative, compared to a parameter estimation problem where it is not informative \citep{SafePriors,Martino25fake}. Therefore, the use of informative (but objective) priors rather than uniform priors on the parameters should be prioritized.

Several strategies are followed to construct data-driven priors in the literature but the main aim should be to guarantee a large concentration rate of the posterior distribution. Whether it is best to use the entire data set to construct the prior or to use a train-then-validate approach depends on the specific application \citep{Martin2019}. Using informative priors also has the advantage of reducing the parameter space for {Bayesian inference computational tools}. As a result, the problem of dimensionality can be mitigated. One potential limitation of data-driven prior construction is that it can be sensitive to the quality and quantity of the data used to construct the prior. In addition, if the data used to construct the prior are not representative of the data used for inference, the resulting inference may be biased or unreliable.

Periodicity is a common property of many astrophysical processes, which are often described by well-known mathematical models. Bayesian inference is frequently used to estimate parameters of those models, whose output is extremely sensitive to changes in the period. Quite often, the period shows a posterior distribution that is highly concentrated in a small subset of the parameter space. Thus, exploring a large space is not efficient and it is quite usual to restrict the support of the prior. In particular, frequency analysis is widely used to reveal dominant frequencies in time series. The information provided by periodograms and scalograms is used in many works to construct informative priors that ensure exploring regions of high probability mass \citep{bretthorst2001,gregory2010,morales2019,lopez2020}. Nevertheless, these priors are arbitrarily chosen to contain the dominant periods. It is common to construct a uniform prior around an intense peak in the periodogram, with support on an interval which is either long enough to ensure that the inference algorithm has some margin of adaptability, or very narrow to mitigate the {curse} of dimensionality. It is important to remark that (a) the length of the support interval is arbitrary and (b) even with a long interval for support, a uniform prior is strongly informative because it forces zero probability mass in regions where the likelihood may be positive \citep{bernardo1985,SafePriors}. Furthermore, effects such as spectral leakage are often not taken into account.

Using the same data for both constructing the prior and performing the inference can introduce biases and invalidate the results, as it would essentially be "double-counting" the information in the data. However, this double use of the data is a standard procedure in the so-called empirical Bayes approach and in cross-validation schemes \citep[e.g][]{bernardo1985,bishop2007,SafePriors}. Moreover, the case of periodic time series is very specific. If the period is stationary, i.e. it does not change with time or vary very little, the information contained in any time series of that process is essentially equivalent. Therefore, using the same data for constructing a prior over the period and for inference is justified.

In this paper, we propose to use a Gaussian process (GP) to construct the prior distribution of the period of a physical model (which can later be used by any Bayesian method to estimate all the unknown model parameters). In particular, we introduce a scheme where a GP endowed with a periodic kernel is fitted using the available data. The GP depends on several hyperparameters and a Bayesian estimation method can provide a posterior probability distribution for each of them. Specifically, one of these hyperparameters is related directly to the periodicity of the GP and its distribution {(computed via importance sampling)} naturally yields a prior distribution for the period of the physical model. The prior distributions over the kernel hyperparameters are constructed based on the periodogram. Our methodology is implemented using importance sampling schemes and tested against both simulated and real-world quasi-periodic time series.
{Therefore, our workflow is, essentially, empirical Bayes. We learn an informative prior for the period by a periodic-kernel GP and, in a modular (\emph{cut}) scheme, pass its posterior forward as a \emph{posterior-as-prior} with no feedback. Equivalently, this modular posterior-as-prior can be considered a \emph{power prior} \citep[see][]{Ibrahim2015} built from the periodic–GP likelihood, with a power $\alpha \in [0,1]$ controlling the degree of borrowing (we use $\alpha=1$ by default). It is worthnoting that, in principle, the posterior-as-prior construction may be affected by the \emph{false confidence theorem} \citep[e.g.][]{martin2019false}, which states that probabilistic belief functions can assign high probability to false parameter values with non-negligible frequency \citep{balch2019satellite,carmichael2018exposition}. In this application, however, the non-parametric posterior for the period parameter closely reflects the underlying periodicity, and thus the potential impact of false confidence on the subsequent parametric inference is expected to be limited.}

GP regression is a well-known method applied to Astronomical data. It has been used, for example, to study the correlation in the noise of radial velocity data \citep[e.g.][]{Affer2019,stock2020,Kossakowski2022}, to estimate stellar parameters \citep{bu2015,bu2020}, to model mid-term stellar variability \citep{colombo2022} and to determine the Hubble constant based on a non-parametric model \citep{liao2019}. Also, GP regression is a popular method for time series analysis \citep[see][for a review]{aigrain2023}. Here we show that the combination of GP regression and importance sampling methods is a powerful tool to construct informative priors {for inference problems} on periodic models. In particular, the use of informative priors can mitigate the curse of dimensionality. It also avoids penalizing models for a model selection problem, as we mentioned above.

\section{Prior construction over the period}
\label{sec:priorConstruction}

\subsection{Concept}

In the literature, periodograms are often used to identify periodic components within time series data. The use of periodograms facilitates the imposition of constraints on the prior distribution of the period parameter when applying Bayesian inference techniques \citep[e.g.,][]{morales2019,MNRAS}. While this methodology provides insight into the period value within a parametric model governing the data generation process, it does not provide a probability distribution suitable for use in inference methods (commonly Markov Chain Monte Carlo, MCMC, or importance sampling, IS). Typically, narrow uniform priors centered on dominant frequencies are constructed for this purpose. An alternative strategy is to fit a GP to the data using a periodic kernel. The hyperparameter associated with the period in such a kernel is directly related to the underlying periodicity within the data set.
In the aforementioned approaches, the value of the period is determined without a corresponding uncertainty. In a Bayesian approach to the problem, one applies an estimation algorithm that yields the (approximate) probability distribution of the kernel period in the GP. The marginal posterior distribution of this hyperparameter can be used as a prior distribution to infer the model parameters. In this way, we construct a sufficiently informative prior for the period.
%
%

\subsection{Formal derivation}

Consider an observed data vector denoted as ${\bf y}=[y_1,...,y_M]^{\top}$ (see Table~\ref{tab:acronyms} for a definition of the variables used hereinafter). It is a prevalent scenario across various application domains to be confronted with different models, {each based on a different likelihood function.} Consider two such likelihood functions, $\ell_1({\bf y}|\boldsymbol{\lambda})$ and $\ell_2({\bf y}|\boldsymbol{\theta})$, with respect to the same data vector yet contingent upon distinct parameter sets, $\boldsymbol{\lambda} = [\lambda_1,...\lambda_{D_\lambda}]$ and $\boldsymbol{\theta}=[\theta_1,...\theta_{D_\theta}]$, respectively. As an example,  {one likelihood ($\ell_1$)} could be induced by a parametric model (derived by physical knowledge of the specific application) an  {the other likelihood ($\ell_2$)} could be induced by a  non-parametric model. Of particular interest is the case wherein certain components, specifically $\lambda_j$ and $\theta_k$, are  {inter-related} or  {represent} analogous physical and mathematical entities. In this scenario, it becomes feasible to merge insights garnered from one model into the  {structure} of the other.

To illustrate with an example, let us assume $D_\lambda=D_\theta=1$,  {and let $\lambda$ and $\theta$ be two random variables that represent the same real-world magnitude. We then consider two probabilistic models, parametrized by $\lambda$ and $\theta$. In model 1 ($M_1$), we are given a certain likelihood function $\ell_1({\bf y}|\lambda)$ and the prior for the parameter of interest is given by the probability density function (pdf) $p_{0,1}(\lambda)$. Additionally, we have a second model ($M_2$) characterized by the likelihood $\ell_2({\bf y}|\theta)$ and the prior density $p_{0,2}(\theta)$. In this scenario, we can build a posterior distribution for $\theta$ using model $M_2$ and then use this posterior to define the prior $p_{0,1}$ of parameter $\lambda$ in model $M_1$.}

 {To be specific, let us choose a (possibly broad) prior $p_{0,2}$ for the random parameter $\theta$ in model $M_2$ and assume we have the means to compute the posterior pdf for this model, namely,
\begin{equation}
p_2(\theta|{\bf y}) \propto \ell_2({\bf y}|\theta) p_{0,2}(\theta). 
\end{equation}
Then, we use this posterior density of $\theta$ as a prior pdf for $\lambda$, i.e., we {\em define}
\begin{equation}
p_{0,1}(\cdot) := p_2(\cdot|{\bf y}),
\end{equation}
and $\lambda \sim p_{0,1}$ in model $M_1$. This implies, of course, that
\begin{equation}
p_{0,1}(\lambda) \propto \ell_2({\bf y}|\lambda) p_{0,2}(\lambda)
\end{equation}
and, hence, the posterior for model $M_1$ becomes
\begin{equation}
p_1(\lambda | {\bf y}) \propto \ell_1(\lambda|{\bf y}) p_{0,1}(\lambda) \propto \ell_1(\lambda|{\bf y}) \ell_2({\bf y}|\lambda) p_{0,2}(\lambda).
\label{eqM12}
\end{equation}
}

 {Equation~\ref{eqM12} can be interpreted as yet another model ($M_3$) where the prior density coincides with the prior of model $M_2$, hence $p_{0,3}=p_{0,2}$ and the likelihood is the product of the likelihoods in models $M_1$ and $M_2$, i.e., $\ell_3({\bf y}|\cdot) \propto \ell_1({\bf y}|\cdot)\ell_2({\bf y}|\cdot)$. This model yields the same posterior pdf as $M_1$, since $p_3(\lambda|{\bf y}) \propto \ell_3({\bf y}|\lambda) p_{0,3}(\lambda) \propto p_1(\lambda|{\bf y})$ by construction. However, the factorization of the inference process in two steps,
\begin{itemize}
\item one to compute the data-informed prior $p_{0,1}(\lambda) \propto \ell_2({\bf y}|\lambda) p_{0,2}(\lambda)$ 
\item and a second step to compute the posterior $p_1(\lambda|{\bf y}) \propto \ell_1({\bf y}|\lambda) p_{0,1}(\lambda)$,
\end{itemize}
can lead to more flexible and efficient inference methods, since it enables the application of different modeling techniques and estimation algorithms at each step.
} This is important, for example, when a parametric model is particularly sensitive to small changes in one or more parameters. The inference of this kind of parameters is particularly challenging and requires informative priors to perform a proper inference.

 {
One may criticize the double use of data to construct the prior $p_{0,1}(\lambda)$ and the second step of inference. However, Eq.~\ref{eqM12} shows that the procedure is still proper. Furthermore, there are several strategies to control the impact of this type of procedure on the results. A discussion of data-informed priors (with a review of the ideas proposed in the literature) is given by \citet{SafePriors}. One of these ideas is to use a tempered version of $p_2(\cdot|{\bf y})$ in order to construct the data-informed, i.e.,
\begin{equation}
p_{0,1}(\lambda) \propto p_2(\lambda|{\bf y})^\gamma
\end{equation}
which yields
\begin{equation}
p_1(\lambda|{\bf y}) \propto \ell_1({\bf y}|\lambda) \left(
    \ell_2({\bf y}|\lambda) p_{0,2}(\lambda)
\right)^{\gamma},
\end{equation}
with $0< \gamma \leq 1$.
} A possible choice of $\gamma$ is
\begin{equation}
\gamma=\frac{1}{M+1},
\end{equation}
where $M$ is the number of {data points} in the vector ${\bf y}$. This choice comes from the usual hypothesis  of conditional independent observations, which yields likelihoods consisting of products of $M$ terms,
\begin{equation}
    \ell_2({\bf y}|\lambda)=\prod_{m=1}^M \ell_2(y_m|\lambda) 
    \quad\text{and}\quad
    \ell_1({\bf y}|\lambda)=\prod_{m=1}^M \ell_1(y_m|\lambda).
\end{equation}
Another possibility is to consider a data-tempering approach by dividing the observations into a training data subset and a test data subset, as in a standard cross-validation (CV) procedure. 

As an application, in this work, we consider that the first likelihood $\ell_1$ is induced by a physical parametric model and the second likelihood $\ell_2$ is based on a  {GP} regression model (i.e. a non-parametric model). For completeness, we include Sect.~\ref{sec:GP} with the formalism of the GP and details on the kernel used in this work. 


\begin{table}[!t]
\caption{List of key variables and functions.}
\label{tab:acronyms}
\centering
 {
\begin{tabular}{l l}
\hline\hline
Variable/function & Meaning \\
\hline
${\bf y}$ & observations \\
$\boldsymbol{\lambda}$ & a generic parameter set  \\
$\boldsymbol{\theta}$ & a generic parameter set  \\
$D_\lambda$ & dimension of parameter set $\boldsymbol{\lambda}$ \\
$D_\theta$ & dimension of parameter set $\boldsymbol{\theta}$ \\
$\ell_1(\mathbf{y}|\lambda)$ & likelihood for a 1D--parameter $\lambda$ \\
$\ell_2(\mathbf{y}|\theta)$ & likelihood for a 1D--parameter  $\theta$\\
$p_{0,1}(\lambda)$ & prior distribution of the parameter $\lambda$ \\
$p_{0,2}(\theta)$ & prior distribution of the parameter $\theta$ \\
$p_1(\lambda|\mathbf{y})$ & posterior distribution of $\lambda$ \\
$M$ & number of data points \\
$\gamma$ & tempering factor \\
\hline
\end{tabular}
}
\end{table}

\section{Gaussian Process}
\label{sec:GP}

Given a set of $M$  {data points labeled with their corresponding time stamps,} $\mathcal{D}=\{ t_i,y_i\}_{i=1}^M$, we assume the following observation model,

\begin{align}
\label{EqOM}
y_i=f(t_i)+\epsilon_i, \qquad i=1,...,M,
\end{align}

\noindent
where $f(t): \mathbb{R}\rightarrow  \mathbb{R}$ is an unknown function {of the independent variable $t$ (here, time)} and $\epsilon_i \sim \mathcal{N}(\epsilon|0,\sigma_e^2)$ for all $i=1,...,M$.  {The variance of the observation noise, $\sigma_e^2$ is not necessarily known and it can be inferred by optimization}. Furthermore, we assume that $f(t)$ can be represented as a realization of a Gaussian Process (GP). We define a {\it kernel function} $k(t, z): \mathbb{R}^{d_x} \times \mathbb{R}^{d_x} \rightarrow \mathbb{R}$,  {and the} corresponding {\it kernel matrix} $[\mathbf{K}]_{i j}:=k\left(t_{i}, t_{j}\right)$ of dimension $M \times M$ containing all kernel entries. Given a generic {time} input $t$, we also define the {\it kernel vector} as $\mathbf{k}_{t}=\left[k\left(t, t_{1}\right), \ldots, k\left(t, t_{M}\right)\right]^{\top}$ of dimension $M\times 1$.  Finally, we denote $\mathbf{y}=\left[y_1, \ldots, y_M\right]^{\top}$ the  $M \times 1$ vector of observed outputs. Thus, GPs provide a Gaussian predictive density 

\begin{align}
p(y | t,\mathcal{D})=\mathcal{N}\left(y| \mu_{\texttt{GP}}(t), \sigma_{\texttt{GP}}^{2}(t)\right),
\end{align}

\noindent
with predictive mean $\mu_{\texttt{GP}}(t)$ and variance $\sigma_{\texttt{GP}}^{2}(t)$. If we assume zero mean and a kernel function $k\left(t, z\right)$ for the GP, then the predictive mean gives us the interpolating function, 

\begin{align}
\mu_{\texttt{GP}}(t)=\widehat{f}(t)&=\mathbf{k}_{t}^{\top} (\mathbf{K}+\sigma_e^2 {\bf I}_M)^{-1} {\bf y}, \\ 
&=\mathbf{k}_{t}^{\top} \boldsymbol{\alpha}, \\
&=\sum_{i=1}^{M} \alpha_{i} k\left(t, t_{i}\right), 
\end{align}

\noindent
where $\widehat{f}(t)=\mu_{\texttt{GP}}(t)$ is an approximation of the unknown function $f(t)$ after observing the data $\mathcal{D}=\{t_i,y_i\}_{i=1}^M$, ${\bf I}_M$ is the $M\times M$ identity matrix, and the weight vector $\boldsymbol{\alpha}=\left[\alpha_{1}, \ldots, \alpha_{M}\right]^{\top}$ is given by $\boldsymbol{\alpha}=(\mathbf{K}+\sigma_e^2 {\bf I}_M)^{-1} \mathbf{y}$,  {where $\sigma_e^2$ is the variance of the data}. 

The GP formulation also provides the expression for the predictive variance,

\begin{align}\label{SigmaGP}
\sigma_{\texttt{GP}}^{2}(t)=k(t, t)-\mathbf{k}_{t}^{\top} (\mathbf{K}+\sigma_e^2 {\bf I}_M)^{-1} \mathbf{k}_{t},
\end{align}

\noindent
{where the first term} is the prior variance $k(t, t)$, and we subtract the second term which is a positive value $\mathbf{k}_{t}^{\top} (\mathbf{K}+\sigma_e^2 {\bf I}_M)^{-1} \mathbf{k}_{t}$, representing the information obtained by the data. Note that the variance is not {\it directly} dependent {on} the observed outputs ${\bf y}$.

\subsection{Kernel function}

The kernel function should encode prior knowledge or belief regarding the smoothness, correlation, and periodicity of the data. Several periodic kernels exist in the literature, from the simple cosine kernel to the exponential sine squared kernel. Of particular interest are the sparse spectrum kernels \citep{lazaro2010}, a.k.a. spectral kernels. They can encode the spectral components of the data. As a result, spectral kernels are useful for data with more than one spectral component. 

For this work and for simplicity, we have chosen the exponential sine squared kernel,

\begin{equation}
\label{eq:kernel}
k_\mathrm{PER}(t, z) = A \exp \left(-\frac{2\sin \left( \frac{\pi}{P} |t-z| \right)^{2} }{ L}\right),
\end{equation}

\noindent
which has 3 hyper-parameters $[A,P,L]$, so that the complete vector of the hyper-parameters of the GP is ${\bm \theta}=[A,P,L,\sigma_e]$. Note that $A = k_\mathrm{PER}(t,t)$ represents the assumed prior variance. We are particularly interested in studying the marginal posterior of $P$ given the data $\mathbf{y}$ (it has the physical meaning of main period of the signal) and the marginal posterior of $\sigma_e$ given the data ${\bf y}$. 

For the application to real data, the previous kernel has been combined with the squared exponential kernel, which is able to model functions that are smooth and infinitely differentiable,

 \begin{equation}
\label{eq:RBF}
k_\mathrm{SE}(t, z) = \exp \left(-\frac{|t-z|^{2} }{ 2l^2}\right).
\end{equation}

In particular, the following kernel has been used for the real data applications, 

\begin{equation}
\label{eq:combinedKernel}
k(t, z) = k_\mathrm{PER}(t, z) \cdot k_\mathrm{SE}(t, z),
\end{equation}

which is able to model quasi-periodic functions. 

\subsection{Tuning of the hyper-parameters}

A typical approach for learning ${\bm \theta}$ is maximizing the so-called marginal likelihood,  {commonly referred to as $p({\bf y})$, i.e. the integral of the product of the likelihood and the prior over the parameter space} \citep[see][]{SafePriors,LlorenteRev}.  {However, this approach does only yield point estimates. Here, we use a Bayesian approach in order to obtain approximations of the marginal posterior densities of the hyper-parameters.} The marginal likelihood is  analytically known

\begin{align}
\ell_2({\bf y}|{\bm \theta})&=\mathcal{N}\left({\bf y}|{\bf 0}, {\bf K}+\sigma_e^2 {\bf I}_M\right), \nonumber \\
&=\mathcal{N}\left({\bf y}|{\bf 0}, {\bf K}_{\texttt{tot}}\right),
\end{align}

\noindent
where we have denoted ${\bf K}_{\texttt{tot}}={\bf K}+\sigma_e^2 {\bf I}_M$. Note that ${\bf K}_{\texttt{tot}}$ depends on the choice of the hyper-parameter vector ${\bm \theta}$. Choosing a prior density over ${\bm \theta}$, denoted as  {$p_{0,2}({\bm \theta})$}, the complete posterior density is given by
 {
\begin{align}
p_2({\bm \theta}|{\bf y})&\propto \ell_2({\bf y}|{\bm \theta})p_{0,2}({\bm \theta}) \nonumber\\
&\propto\mathcal{N}\left({\bf y}|{\bf 0}, {\bf K}_{\texttt{tot}}\right)p({\bm \theta}).
\end{align}
}

We can approximate  {$p_2({\bm \theta}|{\bf y})$} by Monte Carlo sampling methods \citep{Bugallo2017,MartinoLibro} such as importance sampling \citep{Bugallo2017,DeepISFernando, MNRAS}. Recall that, for example in our simulations, we have ${\bm \theta}=[A,P,L,\sigma_e]$. If we consider independent priors, 
 {
\begin{equation}
p_{0,2}(\bm\theta) = p_{0,2}(A,P,L,\sigma_e)=p_{0,2}(A) p_{0,2}(P) p_{0,2}(L) p_{0,2}(\sigma_e),
\end{equation}
(with some abuse of notation) then we have 
\begin{align}
p_{0,2}(A,P,L,\sigma_e|{\bf y}) &\propto \ell_2({\bf y}|A,P,L,\sigma_e)p_{0,2}(A,P,L,\sigma_e), \\
&\propto \ell_2({\bf y}|A,P,L,\sigma_e)p_{0,2}(A) p_{0,2}(P) p_{0,2}(L) p_{0,2}(\sigma_e),
\end{align}
}and the marginal posteriors of the period $P$ and $\sigma_e$ are given by  {
\begin{align}\label{MargPostofP}
p_2(P|{\bf y})
&\propto p_{0,2}(P) \int_{\mathbb{R}^3} \ell_2({\bf y}|A,P,L,\sigma_e )\mbox{ } p_{0,2}(A) \mbox{ }p_{0,2}(L) \mbox{ }p_{0,2}(\sigma_e) \mbox{ }  dA \mbox{ } dL \mbox{ }  d\sigma_e, \\
p_2(\sigma_e|{\bf y})
&\propto p_{0,2}(\sigma_e) \int_{\mathbb{R}^3} \ell_2({\bf y}|A,P,L,\sigma_e )\mbox{ } p_{0,2}(A) \mbox{ }p_{0,2}(L) \mbox{ }p_{0,2}(P) \mbox{ }  dA \mbox{ } dL \mbox{ }  dP,
\end{align}
} that can be both approximated by a Monte Carlo procedure. 

\section{Applications}
\label{sec:applications}


In this study, we conduct a series of experiments to evaluate the performance and reliability of the proposed method when applied to different types of datasets. The analysis begins with a controlled synthetic case: a single sinusoidal signal with additive noise. This scenario allows us to examine the algorithm's ability to detect and characterize periodic features in a simplified context with well-understood properties. {The number of synthetic observations is varied randomly to show the algorithm's behavior when the information provided by the observations is varied.}
Following this initial validation, we apply the method to two real-world astrophysical datasets. The first dataset corresponds to the light curve of a likely binary star system, which exhibits periodic brightness variations due to mutual occultations of the stellar components. The second dataset consists of radial velocity measurements of a star known to host an exoplanet.
To assess the method’s performance, we compare its output with results obtained using traditional frequency analysis techniques, such as the Fourier transform or Lomb-Scargle periodograms. This comparison enables us to quantify the advantages and limitations of our approach in terms of accuracy and robustness.

\subsection{Simulations}
\label{sec:simul}

\begin{figure}
    \centering
    \includegraphics[width=\columnwidth]{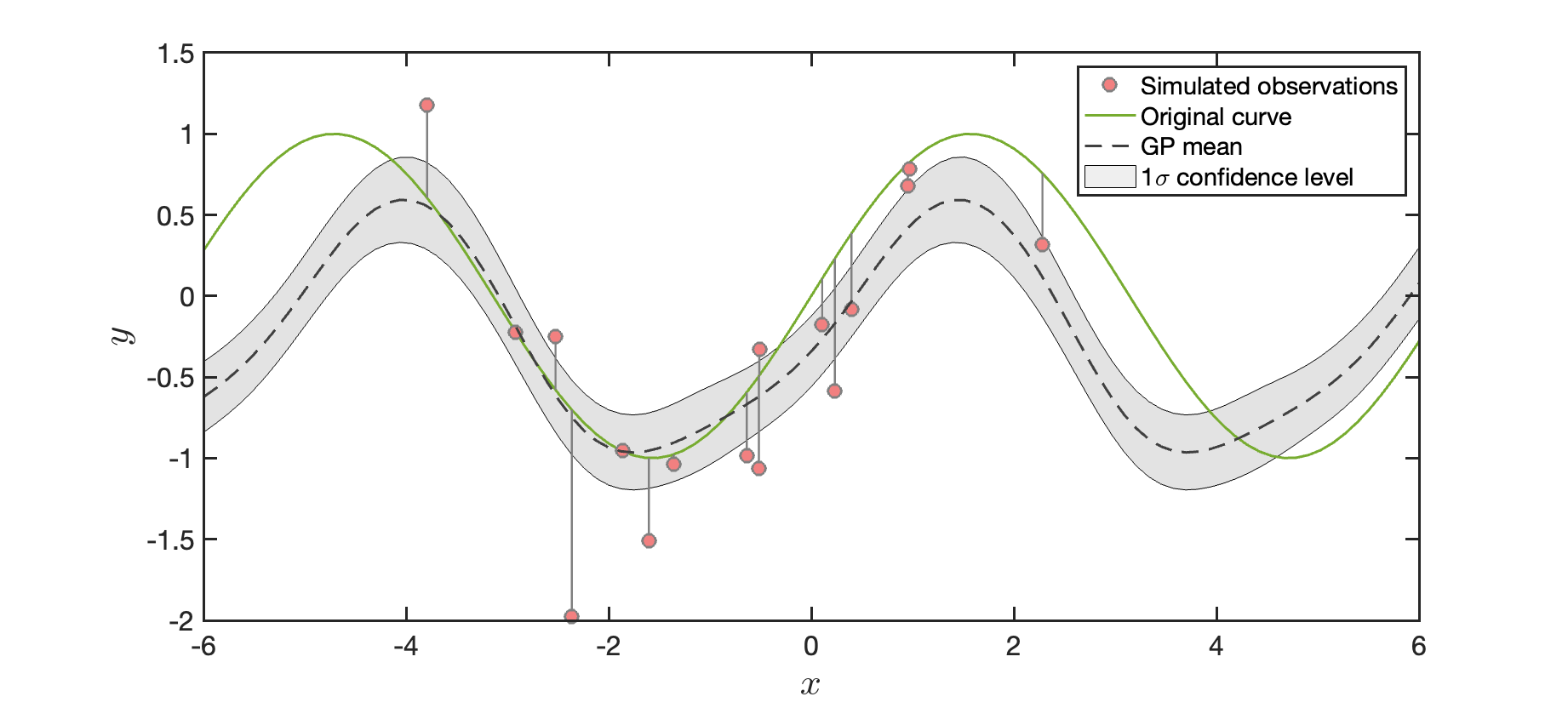}
    \caption{Simulated data (red dots). The green continuous line is the theoretical periodic curve. The result of the GP regression is shown for completeness (black dashed-line).}
    \label{fig:result_simul}
\end{figure}

For the first experiment, we simulated a sinusoidal signal with additive Gaussian noise,
\begin{equation}
    \mathbf{y}(\mathbf{x}) = \sin(\mathbf{x}) + \mathbf{e},
    \label{fig:simulation}
\end{equation}
where $\mathbf{e}$ is a normally distributed  random variable with standard deviation $\sigma_e$ and zero mean, i.e. $p(\mathbf{e}) = \mathcal{N}(0,\sigma_e^2)$. We chose $\sigma_e = 0.5$ for our test. Then, we randomly selected a number of points $x \in [-4,4]$. {Specifically, we simulated 2000 sine waves with the aforementioned parameters and a distinct number of points, from 10 to 100. Figure~\ref{fig:result_simul} shows one example for 16 points (filled circles). For clarity, in the figure} we also plot the sine curve as a continuous line and connect the observations to the line so it is easier to see the deviation of the simulated observations from the theoretical curve. 

We previously mentioned that the posterior distribution for periodic data {is often} concentrated in a small region of the parameter space, so there is a need to use informative priors to avoid dimensionality issues when inferring model parameters. The key parameter in our toy model is the period, $P$. Our goal is to construct an informative prior distribution for $P$ making inference on the GP hyperparametres of a periodic kernel (Eq.~\ref{eq:kernel}).

\begin{figure}
    \centering
    \includegraphics[width=\columnwidth]{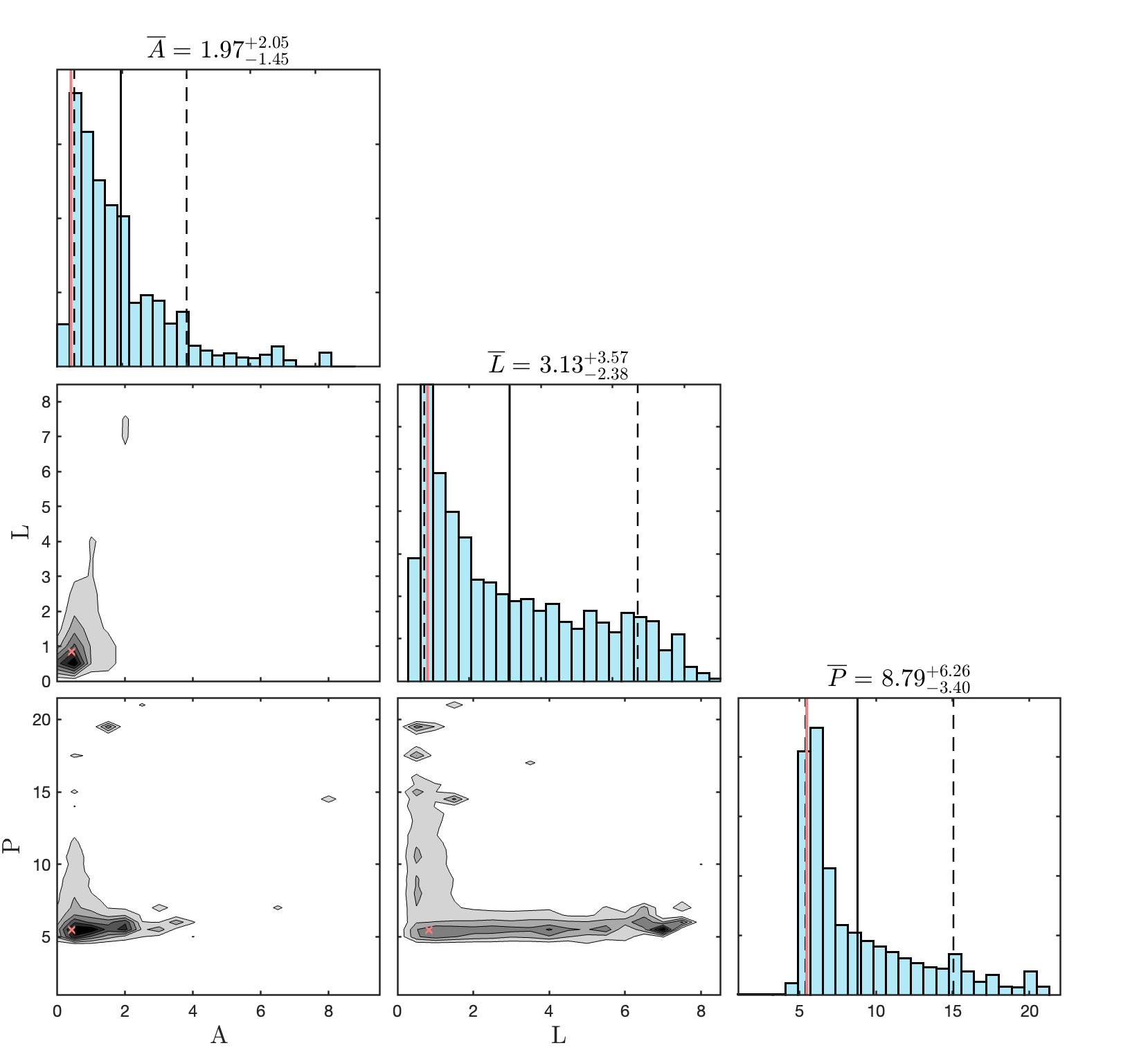}
    \caption{Marginal posterior distribution for the hyperparametres $A$, $L$ and $P$. The value over each histogram is the mean and the 90\% credibility level (continuous black line and dashed lines respectively). The red continuous line represents the estimated MAP.}
    \label{fig:cornerPlot}
\end{figure}

An adaptive importance sampling (AIS) scheme was used to sample from a proposal distribution function.  {Algorithm~\ref{al:AIS} shows a generic implementation of the AIS \citep[e.g.][]{Bugallo2017}. Note that, in this algorithm, the target distribution $\pi(\boldsymbol{\theta}|\mathbf{y})$ is the non-normalized proposal, i.e. $\pi(\boldsymbol{\theta}|\mathbf{y}) = p(\boldsymbol{\theta}|\mathbf{y}) p(\boldsymbol{\theta})$, where $p(\boldsymbol{\theta}|\mathbf{y})$ is a likelihood function. Any importance sampling (IS) method relies on sampling from a proposal distribution that needs to have a support that covers the posterior distribution. If the algorithm contains an adaptation scheme, then, this proposal is iteratively adapted in order to better approximate the posterior distribution at each iteration step. The adaptation seeks the construction of a proposal distribution with a support that is similar to that of the posterior distribution. For the sake of simplicity, we used a normal distribution for the proposal distribution, with initial large variance.} The GP kernel has three hyperparametres, named $A$, $L$ and $P$ (for simplicity, $\sigma_e$ is fixed to the value chosen for the simulation). At each step of the AIS, we adapted the proposal mean and variance. After a number of iterations, we derived the marginal posterior distribution of $A$, $L$ and $P$, by resampling from the weights of the AIS samples. {In Fig.~\ref{fig:cornerPlot}, we show the results for the example presented in Fig.~\ref{fig:simulation}}. The continuous, black line denotes the mean of the samples. The dashed lines are the 90\% credibility levels. Finally, the continuous, red line is the maximum a posteriori (MAP) of the  {marginal posterior distributions, $A = 0.42$, $L = 0.84$ and $P = 5.47$, respectively.} 
The important result is the approximate marginal probability density functions ({see the} histograms in Fig.~\ref{fig:cornerPlot}). 

\begin{algorithm}
{
\SetAlgoLined 
\tcc{{\bf Initialization:}}
- Choose the number of iterations, $T$\;
- Choose the number of samples per iterations, $N$\;
- {Choose an initial proposal $q(\params|\boldsymbol{\beta}_1)$, where $\boldsymbol{\beta}_1$ is the set of proposal parameters. For example, if $q(\params|\boldsymbol{\beta}_1)$ is a multivariate Gaussian pdf, then $\boldsymbol{\beta}_1 = \{ \boldsymbol{\mu}_1, \boldsymbol{\Sigma}_1 \}$, where $\boldsymbol{\mu}_1$ is the mean vector and $\boldsymbol{\Sigma}_1$ is the covariance matrix of the Gaussian distribution.} \\
 \tcc{{\bf Iterations:}}
 \For{t = 1, ..., T}{
  \tcc{{\bf Sampling:}}
       - {Draw $\params_{1,t}, \ldots, \params_{N,t} \sim q(\params|\boldsymbol{\beta}_t)$}\;
       - Compute the unnormalised weights 
       $$
       w(\params_{i,t})=\frac{\pi(\params_{i,t}|{\bf y} )}{{q(\params_{i,t}}|\boldsymbol{\beta}_t)}.
       $$ 
       for $i=1,..,N$\;
       \tcc{{\bf Adaptation of proposal:}}
         - {Compute a new parameter set $\boldsymbol{\beta}_{t+1}$ for the proposal distribution, using the information of the weighted samples}\;
  }
  \tcc{{\bf Output:}}
  - Return the $NT$ weighted samples. $\{\params_{i,t},w(\params_{i,t})\}_{i=1}^N$ with $t=1,..,T$.
 \caption{Generic Adaptive Importance Sampling}
 \label{al:AIS}
 }
\end{algorithm}

\begin{figure}
    \centering
    \includegraphics[width=\columnwidth]{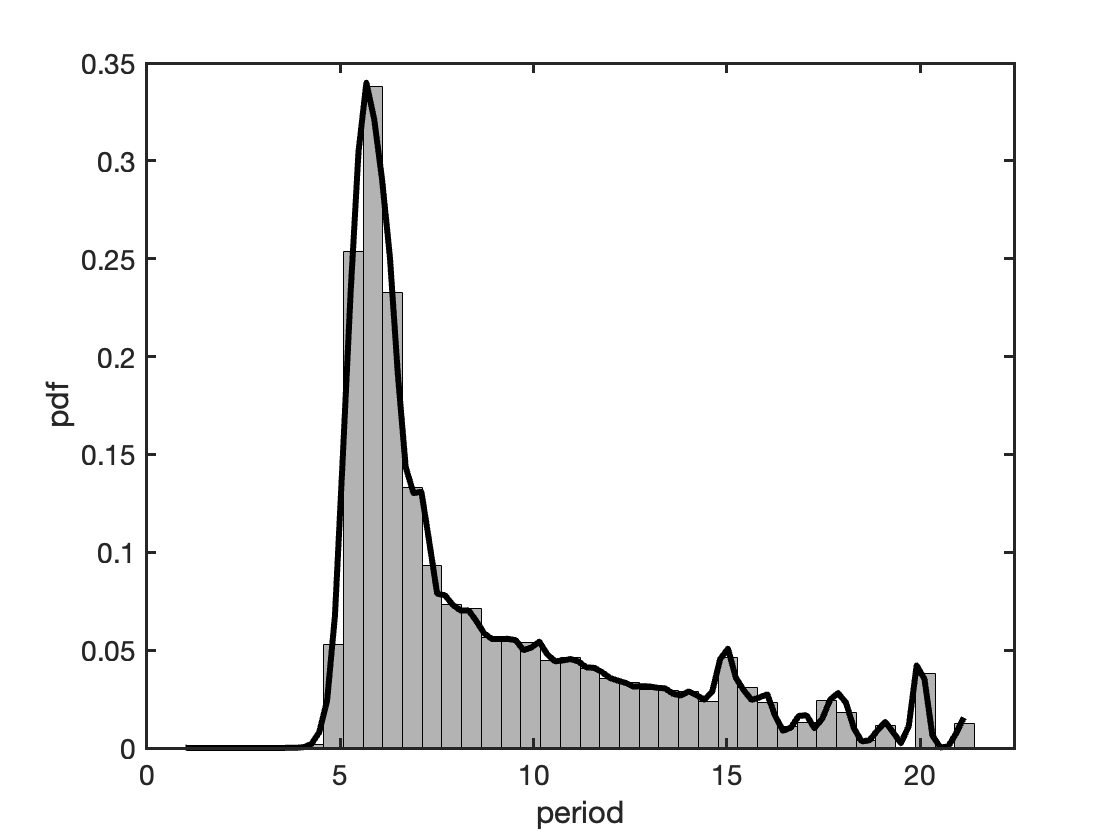}
    \caption{Kernel density estimation (continuous line) for the marginal posterior of the hyperparameter $P$ (histogram).}
    \label{fig:KDE}
\end{figure}

 {The marginal distribution function of the hyperparameter $P$ shown in Fig.~\ref{fig:cornerPlot} (bottom, right panel) can be used as the prior for the period of the generic sine {curve} in Eq.~\ref{eq:modelSimul}. To illustrate the procedure, we first estimated the probability density using a kernel density estimation (KDE). The result is shown in Fig.~\ref{fig:KDE}. We used a normal distribution approximation with optimal bandwidth \citep{bowman1997}. This estimation was used as the prior for the period in the parametric model}

 {
\begin{equation}
    \mathbf{y}(\mathbf{x}) = A\sin \left( \frac{2\pi}{P} \mathbf{x} + \phi \right) + b.
    \label{eq:modelSimul}
\end{equation}
}

\begin{figure}
    \centering
    \includegraphics[width=\columnwidth]{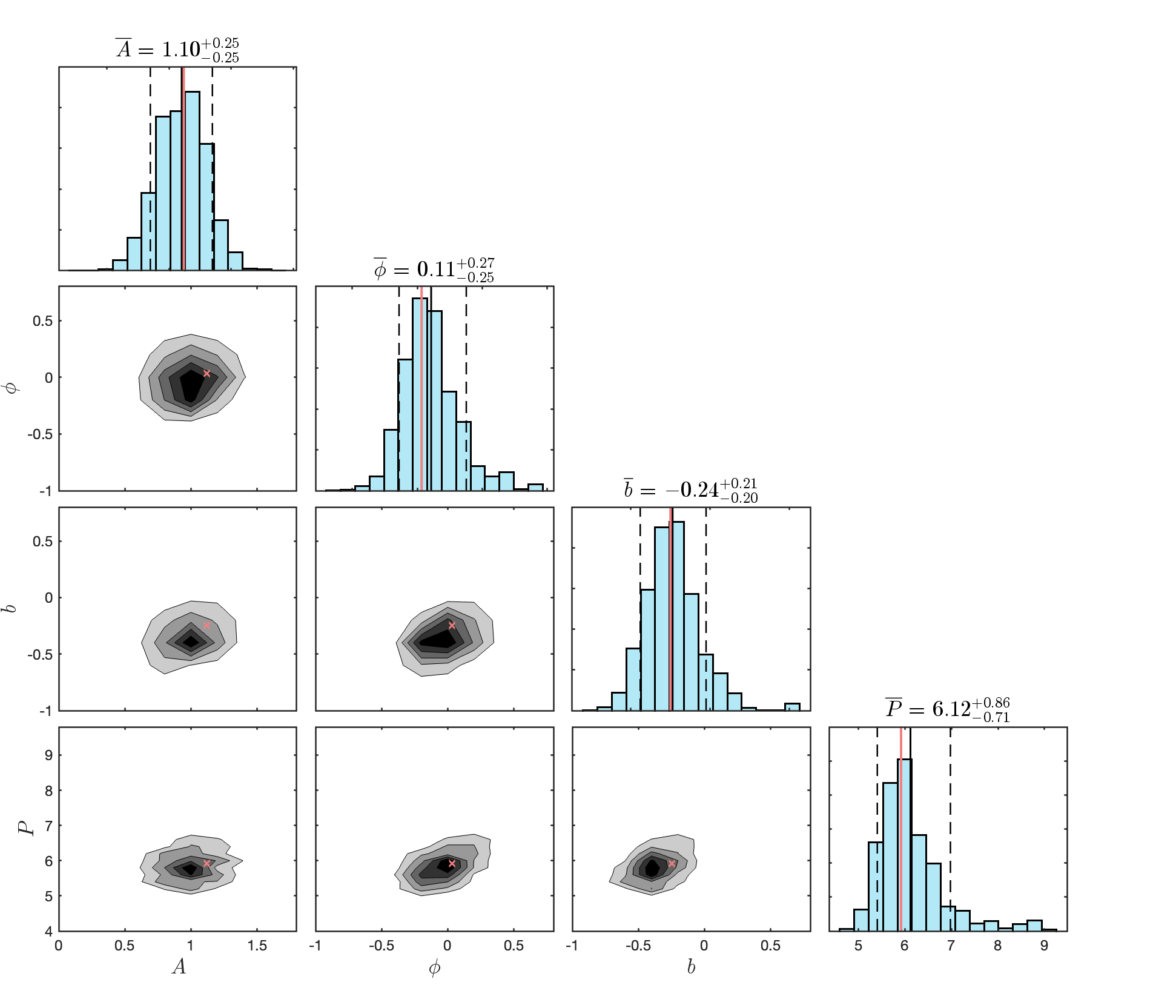}
    \caption{ {Marginal posterior distribution for the model parameters $A$, $\phi$, $b$ and $P$. The value over each histogram is the mean and the 90\% credibility level (continuous black line and dashed lines, respectively). The continuous red line represents the estimated MAP.}}
    \label{fig:cornerPlotModel}
\end{figure}

 {In the following, $P$ is the period of the parametric model. The prior for the remaining parameters in Eq.~\ref{eq:modelSimul} was assumed to be uniform for simplicity. An adaptive importance sampling scheme as used in GP regression (Algorithm~\ref{al:AIS}) was used to approximate the model parameters. Figure~\ref{fig:cornerPlotModel} shows the resulting marginal posterior distributions for each model parameter. Note that the posterior distribution of the period $P$ is remarkably symmetric compared to that used as the prior (see Fig.~\ref{fig:cornerPlot}) and more concentrated. In general, the posterior distribution of the period estimated with the GP regression is wider than that of the parametric model. 
}

\begin{figure}
    \centering
    \includegraphics[width=\columnwidth]{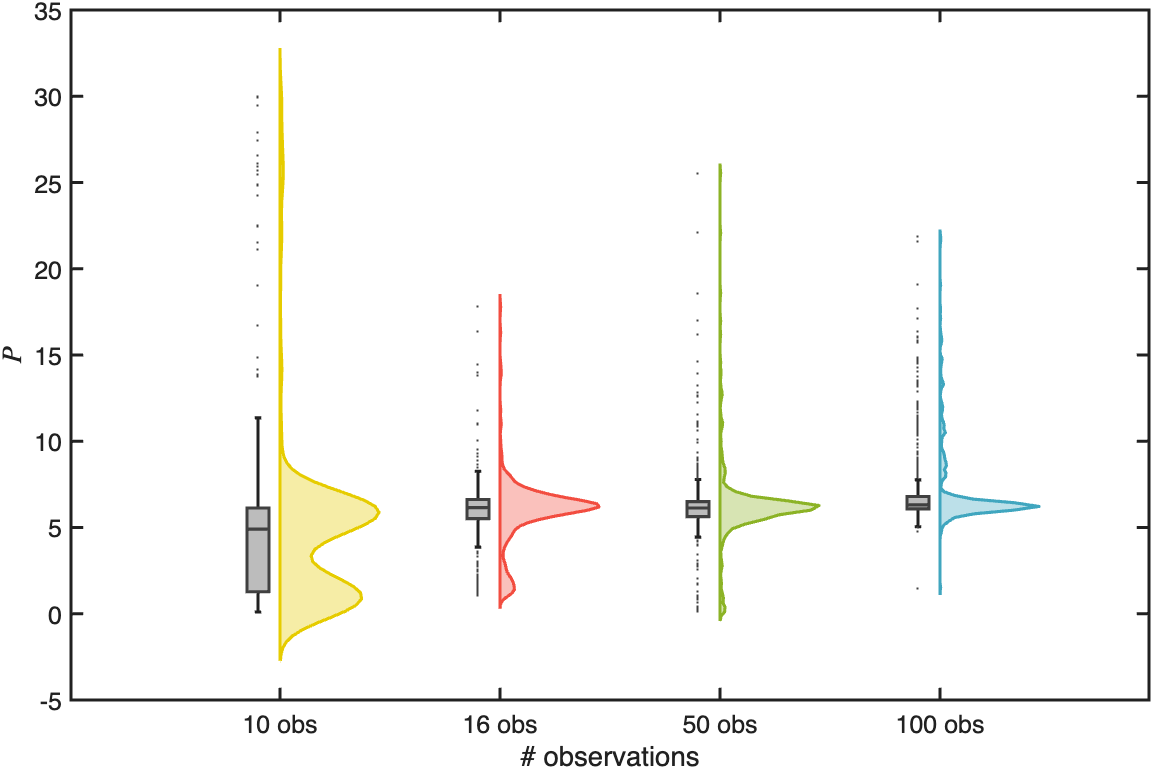}
    \caption{Box chart with the MMSE distribution for the hyperparameter $P$, for simulations with different number of points.}
    \label{fig:violinplot}
\end{figure}

{To demonstrate the stability of the method, we estimated the minimum mean square error (MMSE) for the hyperparameter $P$ in each simulation. Then, we examined its distribution. Figure~\ref{fig:violinplot} shows a box chart of the MMSE, combined with a violin plot. Each box corresponds to simulations with a different number of points. It represents the median and the lower and upper quartiles. The small dots are outliers. The minimum and maximum non-outlier values are shown as error bars. In addition, the violin plot shows the MMSE distribution for the hyperparameter $P$ as obtained from the AIS. As shown in the figure, the MMSE distribution is unimodal, with a maximum close to the simulation value of $P$, and is essentially symmetric for all the simulations except when the number of data points is very low. However, this behavior is expected, as sinusoids with very high frequency can fit well the observations when they are sparse.}

\subsection{Real data}
\label{sec:realdata}

 {We have applied this methodology to real data from two different sources and typologies. For the first case, we chose the radial velocity data from the exoplanet host star GJ 3512 \citep{morales2019}. For the second case, we chose photometric data from the All-Sky Automated Survey \citep[ASAS,][]{pojmanski2004}. We chose a noisy light curve of the binary star HD~112661. With these two examples, we checked the behaviour of the method with two common scenarios: i) extremely concentrated posterior distributions and ii) observations with large uncertainties. More details are given in the following sections. 
}

\subsubsection{Radial velocity curve}
\label{sec:rvData}

For this example, we used data from \cite{morales2019}. The data correspond to radial velocity measurements of the star GJ~3512, which hosts an exoplanetary system.  They were acquired with the CARMENES instrument, a high resolution, optical and near-infrared spectrograph mounted on the 3.5m telescope at the Calar Alto Observatory \citep{CARMENES}.  {The 158 observations expand for 900 days. In the present work, only optical data are utilized, without any loss of generality. The mean value for RV uncertainties is $\sim 2$~m\,s$^{-1}$ for the optical channel of the instrument.} 

\begin{figure}[!t]
    \centering
    \includegraphics[width=\columnwidth]{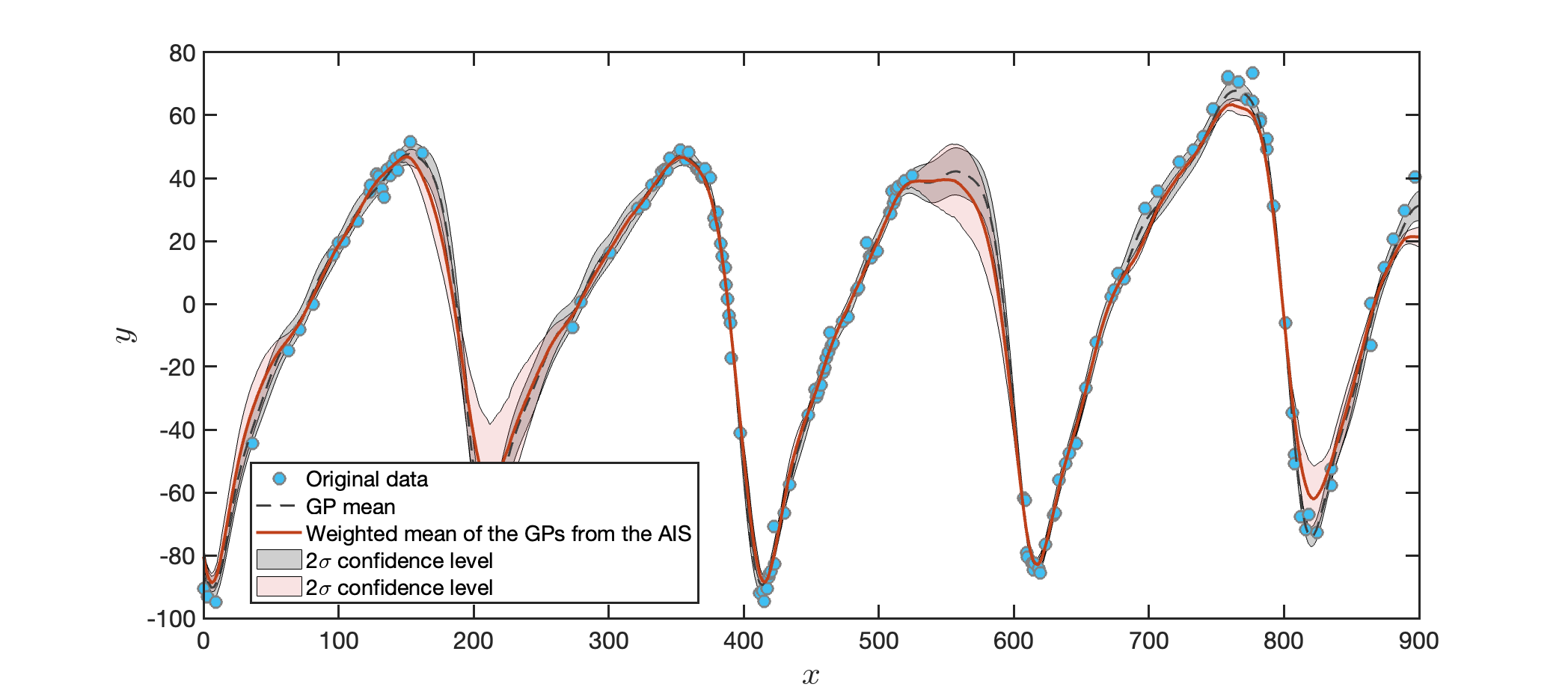}
    \caption{GP regression and weighted mean of the GP from the AIS, with variance \citep[usually referred to as full Bayesian solution][]{Martino2021}. Real data from \cite{morales2019}.}
    \label{fig:FBS_real}
\end{figure}

The observations can be succinctly modeled by

\begin{align}
\label{eq:obs}
y_i = v_i + \xi_i,
\end{align}

\noindent
where $y_i$ is the observation of the star's radial velocity at time $i$, which combines the intrinsic velocity component $v_i$ and an additive noise term $\xi_i$. The parametric model is defined by Eqs.~\ref{eq:obs} and

\begin{align}
\label{eq:explanet}
v_i = V_0 + K \left[ \cos \left(u_i + \omega \right) + e \cos \left(\omega\right)\right], \quad i=1,\ldots,M.
\end{align}

\noindent
where $M$ is the number of observations, $V_0$ is the mean radial velocity of the star, $K$ is the amplitude of the curve, $u$ is the true anomaly, $\omega$ is the periastron angle and $e$ is the orbit eccentricity \citep[see][for a detailed definition of the parametric model]{lopez2020}.
For the standard deviation of the random variable $\boldsymbol{\xi}$, we used $\sigma_s = 3.33$~m\,s$^{-1}$. This value was determined by \cite{lopez2020}.
The kernel we used for this application is a composition of the periodic kernel from Sect.~\ref{sec:simul} and a squared exponential kernel,

\begin{eqnarray}
k(\mathbf{x}, \mathbf{z}|{\bm \theta}) = & \sigma_1^2 \exp \left(-\frac{2\sin \left( (\pi/P) |\mathbf{x}-\mathbf{z}| \right)^{2} }{ l_1^2}\right) + \nonumber \\
 & \sigma_2^2 \exp \left( \frac{ (|\mathbf{x}-\mathbf{z}| )^{2} }{ 2l_2^2} \right),
 \label{eq:myKernel}
\end{eqnarray}

\noindent
where we let the variances of both kernels $\sigma_1^2$ and $\sigma_2^2$ take distinct values, that is, to be different. This kernel allows for small deviations from periodicity. This addition operation in the kernels assure we can capture irregular variability and instrumental drift, contrarily to the multiplicative kernel, which consider changes in the amplitude or shape of the cycles.  
Adaptive importance sampling was applied to the kernel hyperparameters to derive the full Bayesian solution. Figure~\ref{fig:FBS_real} shows the results of the Gaussian process regression and the full Bayesian solution. In general, the means are very similar, but the variance of the full Bayesian solution is larger in some regions. 

\begin{figure*}[!ht]
    \centering
    \includegraphics[width=\columnwidth]{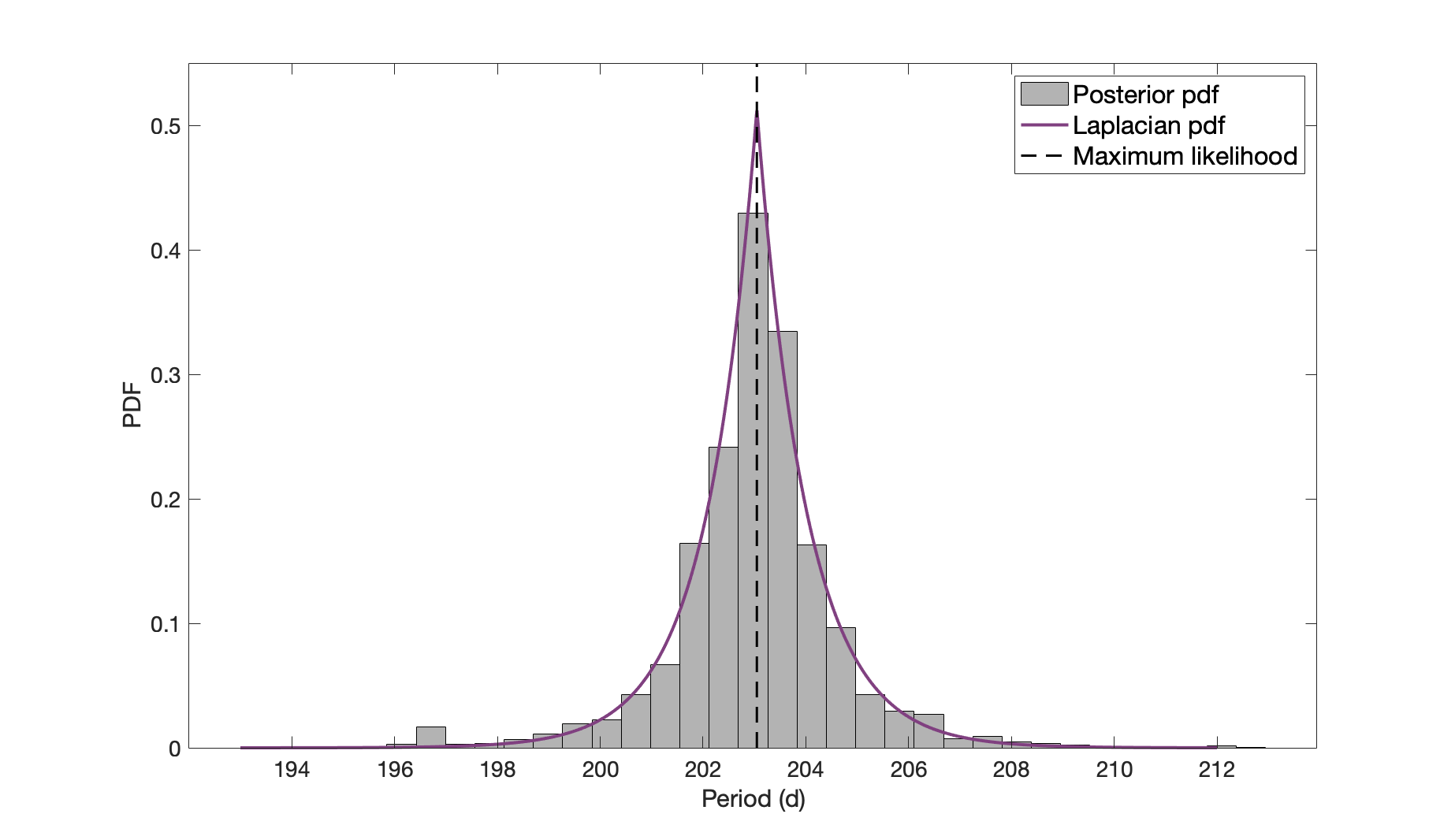}
    \includegraphics[width=\columnwidth]{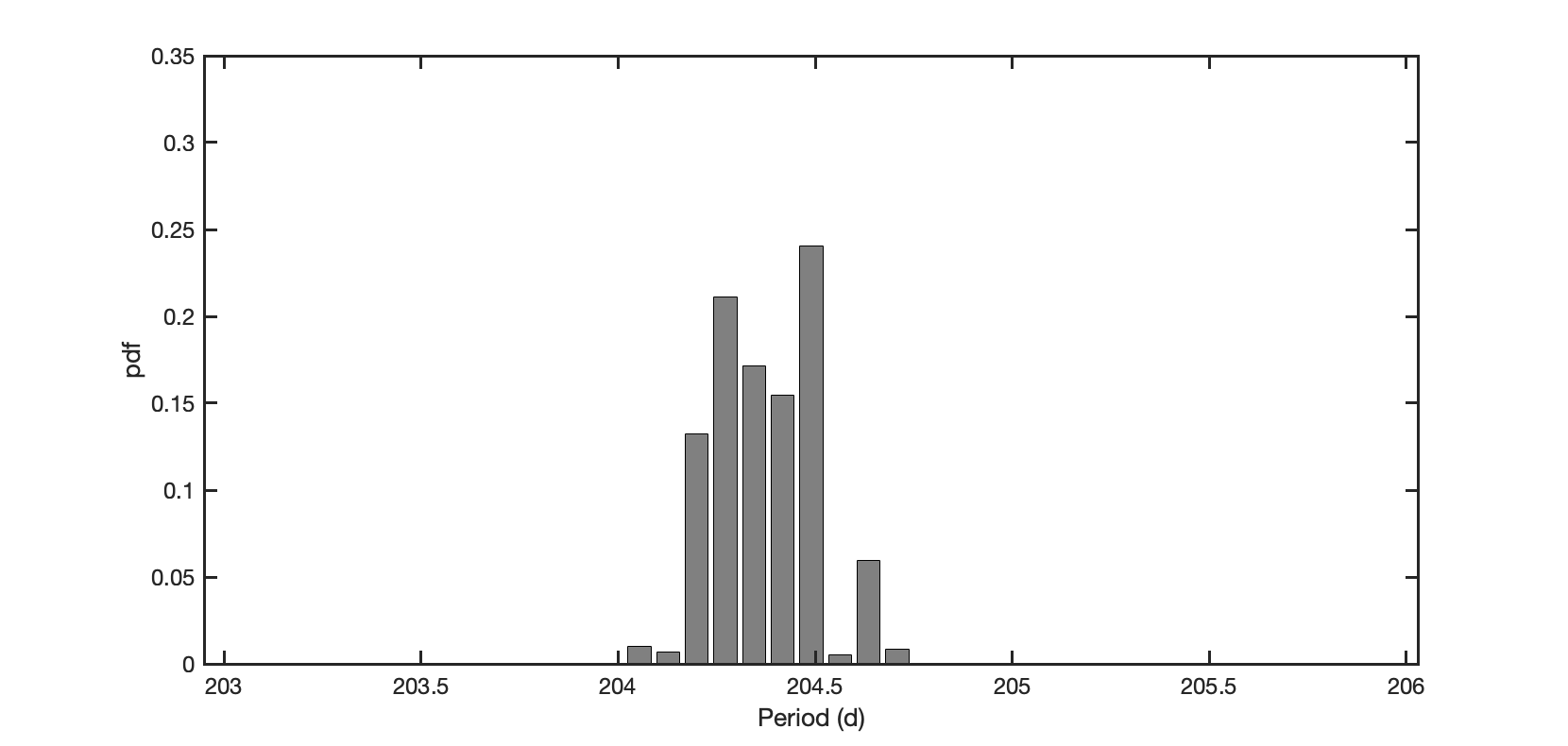}
    \caption{\textbf{Left:} Marginal posterior distribution determined for the hyperparameter $P$ in Eq.~\ref{eq:myKernel}. \textbf{Right:} Marginal posterior distribution of the period in the model obtained from Bayesian inference using the distribution in the left panel as a prior.}
    \label{fig:posteriorP_real}
\end{figure*}

In Sect.~\ref{sec:simul}, we identified the hyperparameter $P$ as the equivalent of the period in the model. It is noteworthy that this period is encompassed within the parameter $u$ in Eq.~\ref{eq:explanet}. Our objective was to formulate a prior for this parameter supporting the comprehensive Bayesian solution obtained previously for the hyperparameters of the Gaussian process (GP) kernel. The left panel of Fig.~\ref{fig:posteriorP_real} shows the marginal posterior distribution of the hyperparameter $P$ from the complete Bayesian solution. For simplicity, we applied a Laplacian function fit to this distribution and employed it as the prior probability density function (pdf) for the subsequent inference of the orbital period of the planet within the parametric model.

The right panel of Fig.~\ref{fig:posteriorP_real} shows the marginal posterior distribution approximated through AIS \citep{math9070784,MNRAS}, using the aforementioned Laplacian-derived prior pdf for the orbital period. The histogram reflects a posterior distribution for this parameter {that has support on a very narrow interval}. In contrast to the findings reported in \cite{lopez2020}, our approach facilitated the approximation of the marginal posterior for the orbital period. This result underscores the importance of constraining the prior for the parameter $P$ which contributes to a more accurate estimation of its posterior distribution. {The mean of the posterior distribution is 204.2~d, which can be compared with 204.5~d obtained in \cite{lopez2020} for a single planet model and 203.6~d obtained in \cite{morales2019}. In those works, the authors constructed uniform priors for the period of the model. \cite{lopez2020} used an AIS method for inference, similar to the algorithm we used in this study, while \cite{morales2019} used a Markov-Chain Monte Carlo (MCMC) method.}

\subsubsection{Photometric light curve}

 {For this example, we selected data from the ASAS source id. 125922-6217.3. This source is the counterpart of HD~112661, an evolved B star classified as B0/1 III/IV by \citet{Houk1975} and B2 IVn by \citet{pantaleoni2021}. The observations expand during approximately nine years for a total of 600 data points. The uncertainties in the magnitude are of the same order of the amplitude of the low amplitude variations in the light curve. 
}

\begin{figure*}
    \centering
    \includegraphics[width=\columnwidth]{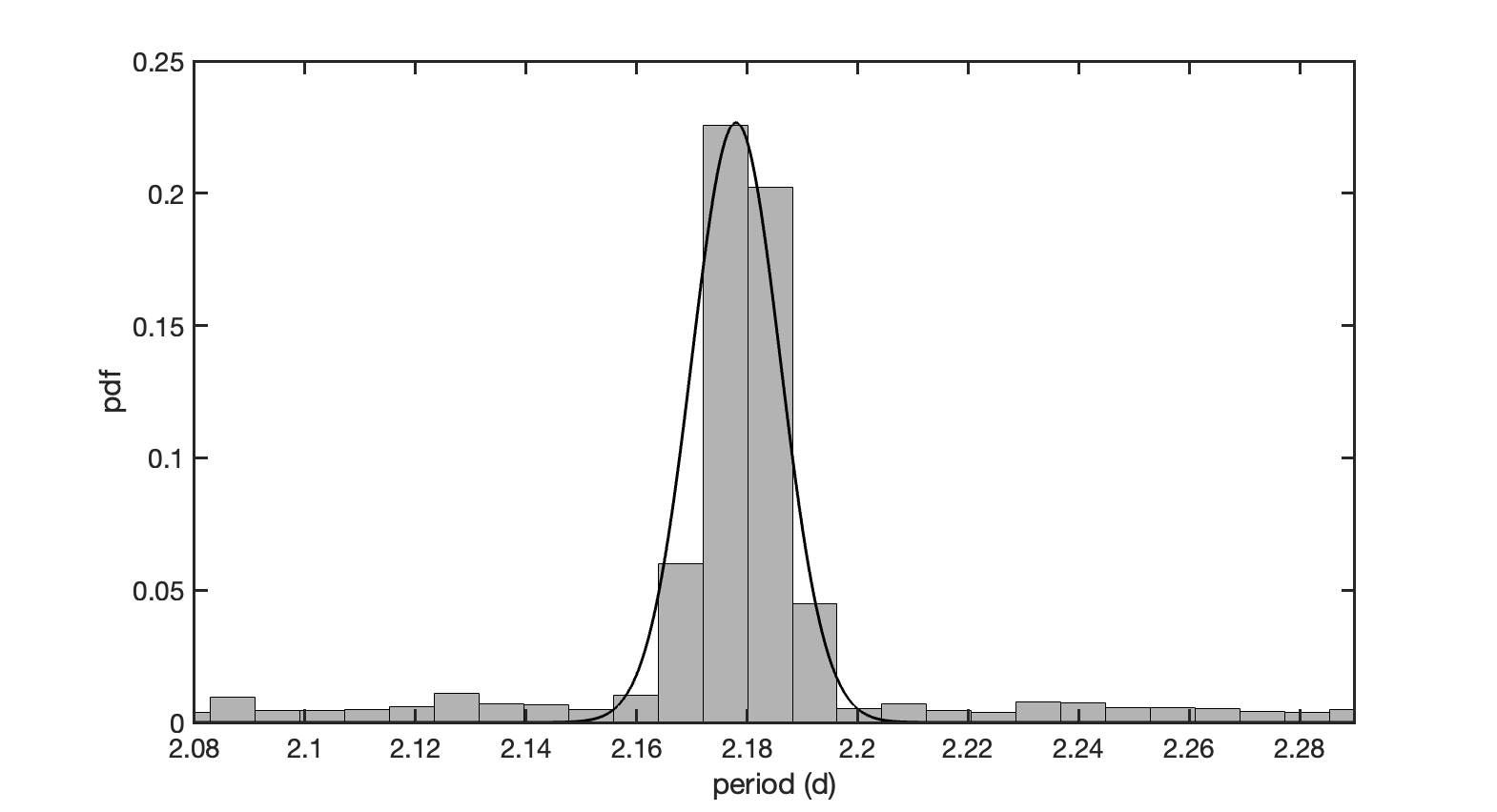}
    \includegraphics[width=\columnwidth]{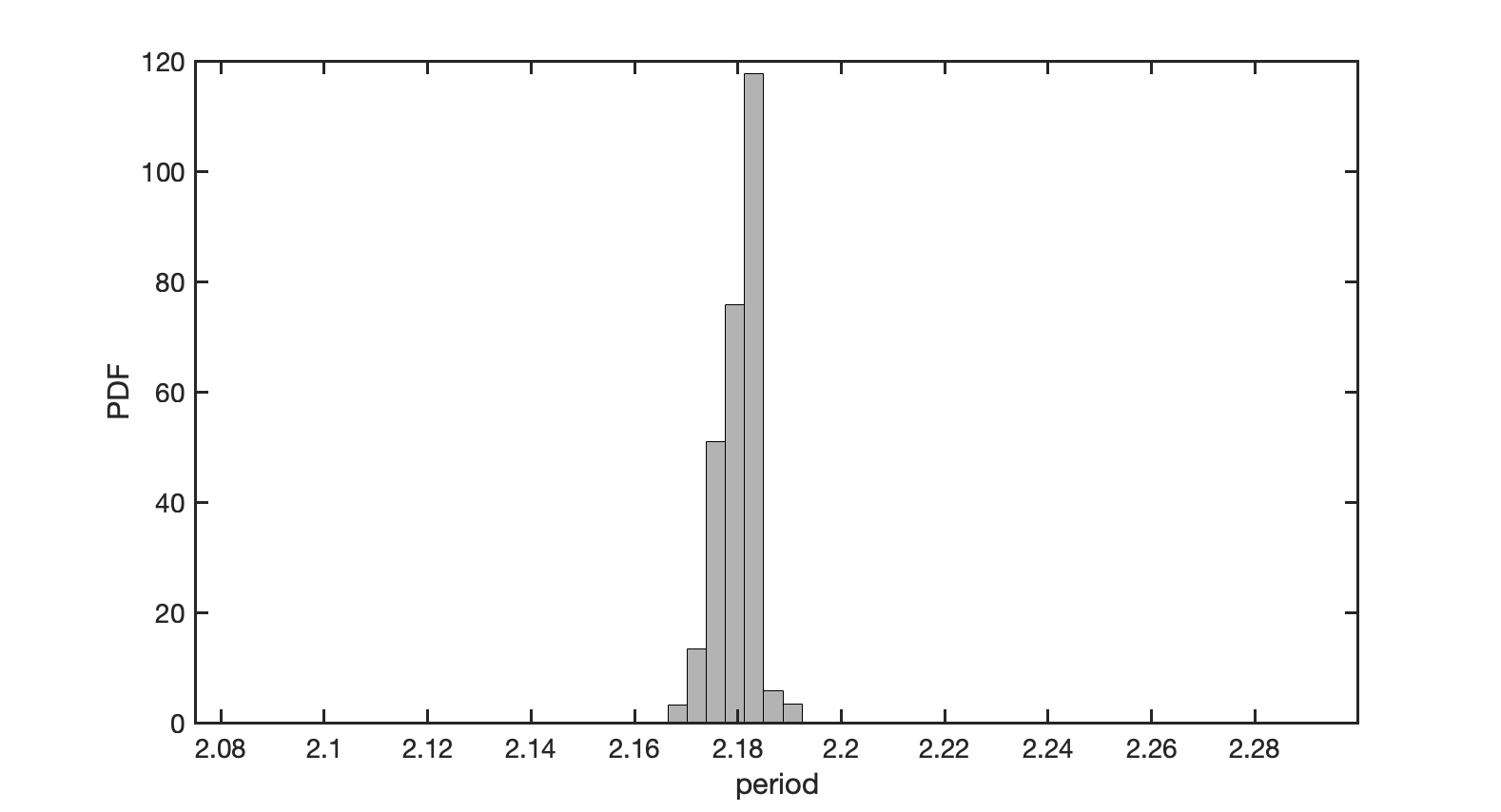}
    \caption{ {\textbf{Left:} Marginal posterior of the hyperparameter $P$ in the GP regression for HD 112661. \textbf{Right:} Marginal posterior of the period in the parametric model.}}
    \label{fig:ASASGPPosterior}
\end{figure*}

 {We followed the procedure described in Sect.~\ref{sec:simul}. First, a GP with a periodic kernel as in Eq.~\ref{eq:kernel} was used to construct the informative prior over the period parameter in the parametric model. This prior distribution was then used for the inversion problem. The parametric model is given in Eq.~\ref{eq:modelSimul}. The left panel in Fig.~\ref{fig:ASASGPPosterior} shows the marginal posterior distribution of the GP kernel hyperparameter $P$. The continuous line is the fit to a normal distribution. We used this normal distribution as the prior of the period of the parametric model in the inversion problem. The right panel shows the marginal posterior approximated for this model parameter in the inversion problem. As in the previous example, the support of the marginal posterior distribution for the period in the parametric model is more concentrated than that of the same kernel hyperparameter.}

\begin{figure}
    \centering
    \includegraphics[width=\columnwidth]{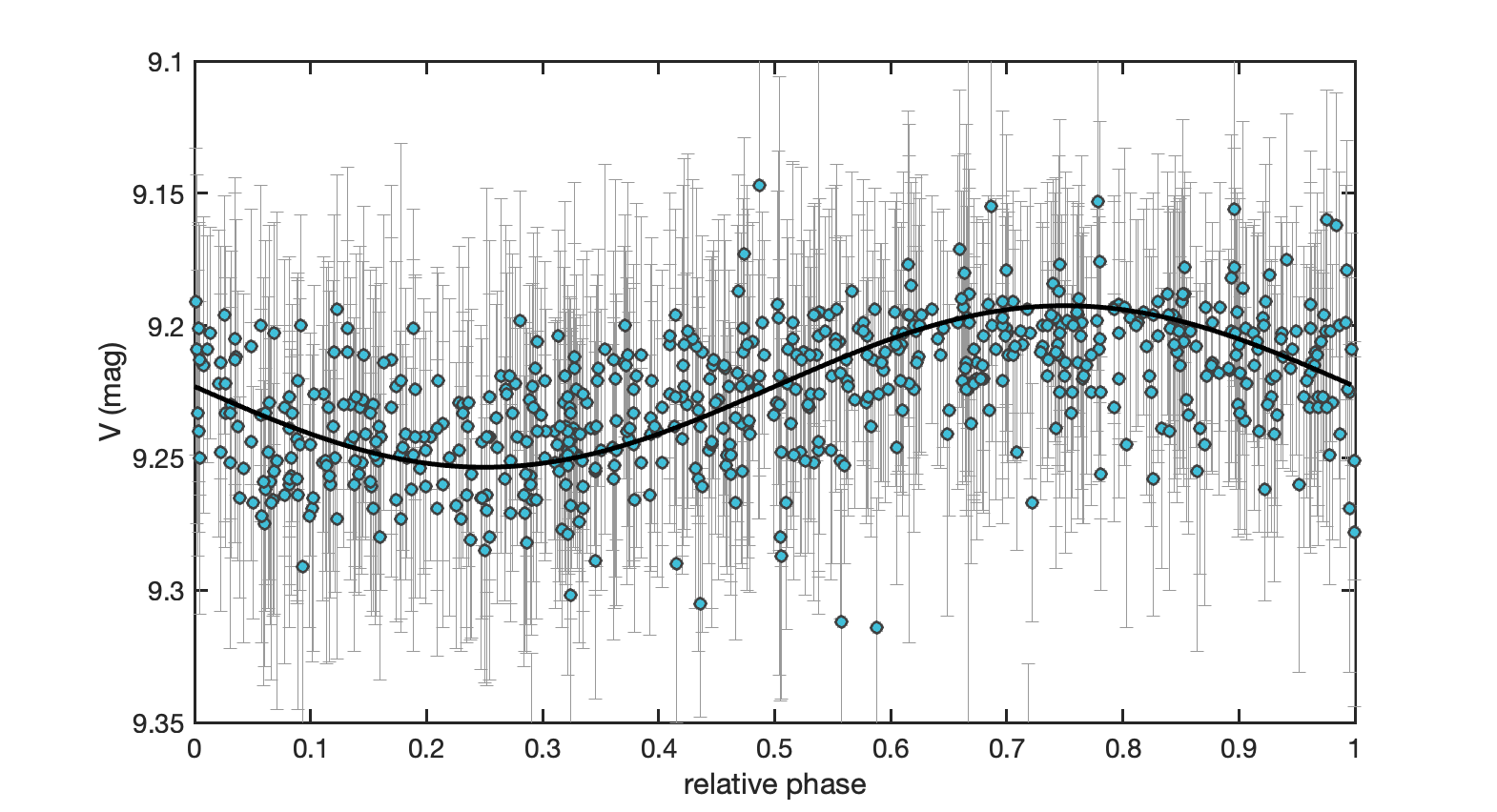}
    \caption{ {Result of the inference for the ASAS data of HD~112661. The light curve is plotted in phase, for clarity.}}
    \label{fig:ASASmodelFit}
\end{figure}

 {The result of the model inference is shown in Fig.~\ref{fig:ASASmodelFit}. The data are plotted in phase for clarity. The estimation for the model parameters is given in Table~\ref{tab:ASASmodelFit}, together with the $90\%$ credibility levels.}

\begin{table}[t]
\caption{Estimated model parameters for the light curve of HD 112661.}
\label{tab:ASASmodelFit}     
\centering                  
\begin{tabular}{c c c c}          
\hline\hline                        
 $A$ & $\phi$ & $b$ & $P$ \\    
\hline                                    $0.03^{+0.01}_{-0.01}$ & $-9.51^{+0.27}_{-0.38}$ & $9.23^{+0.01}_{-0.01}$ & $2.18^{+0.01}_{-0.01}$ \\      
\hline                                      \end{tabular}
\end{table}

\subsection{Comparison with frequency based analysis}

 {In this section, we compare the results obtained using our method with those obtained using a method based purely on frequency analysis. Our aim is to show the performance of GP regression with an AIS against that of frequency analysis in the case of Bayesian inversion (i.e., parameter estimation). For comparison, we used the real data from Sect.~\ref{sec:rvData}.}

\begin{figure}[!t]
    \centering
    \includegraphics[width=\columnwidth]{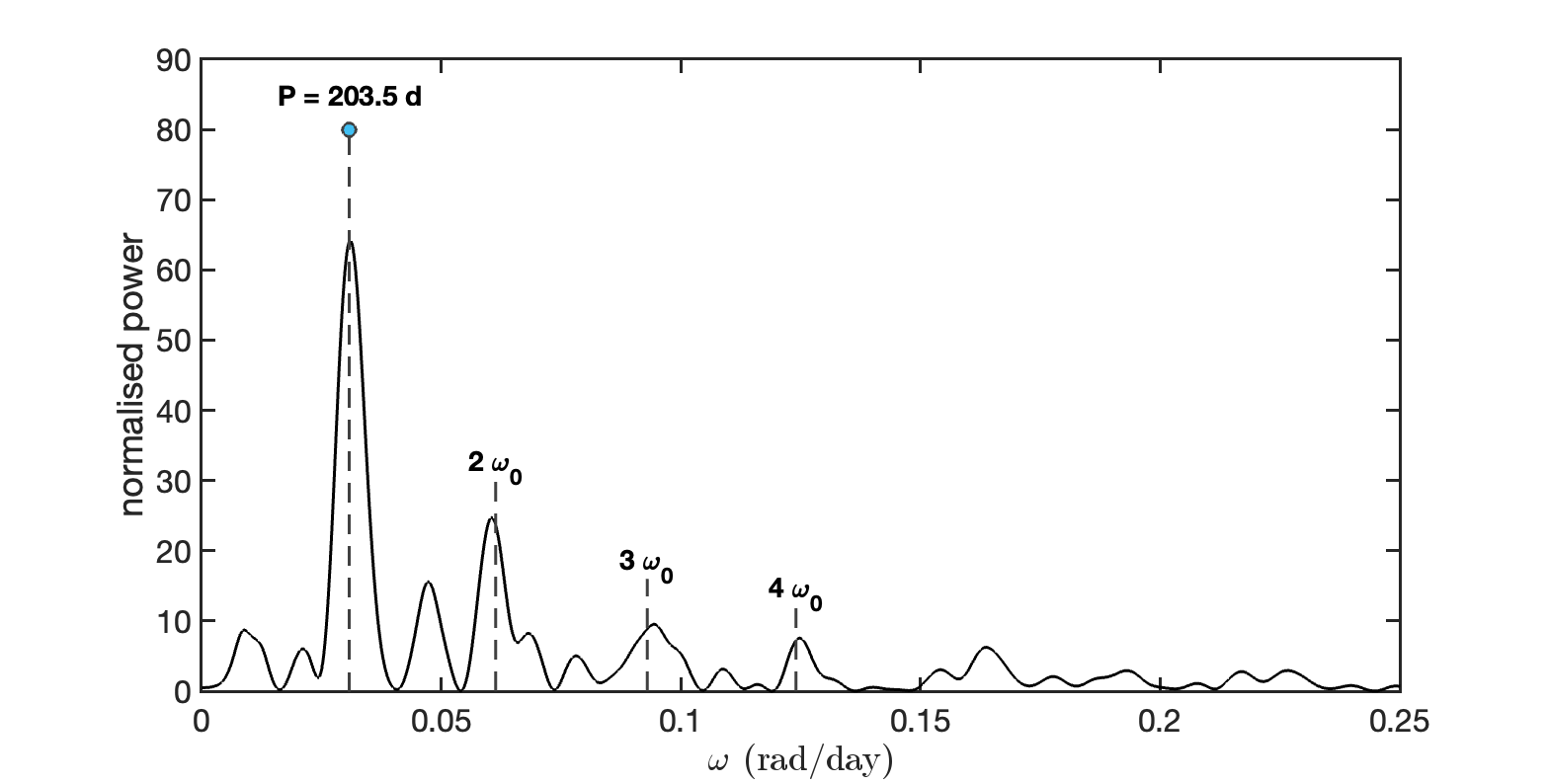}
    \includegraphics[width=\columnwidth]{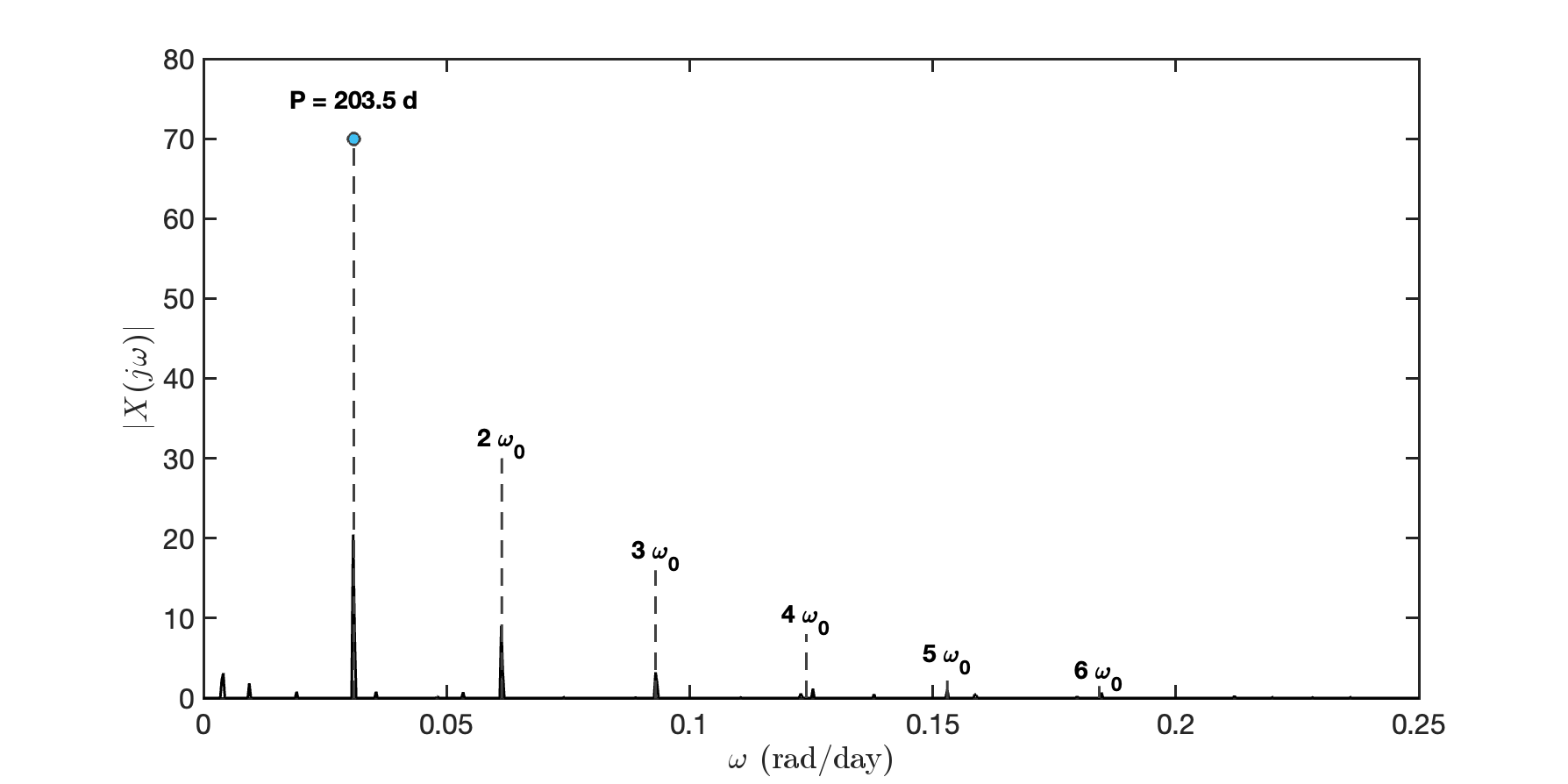}
    \caption{\textbf{Top:} Lomb-Scargle periodogram with the main period at 203.5~d and 3 harmonics identified.  \textbf{Bottom:} L1-periodogram with the main period and 5 harmonics identified.}
    \label{fig:periodograms}
\end{figure}

\begin{figure}[t]
    \centering
    \includegraphics[width=\columnwidth]{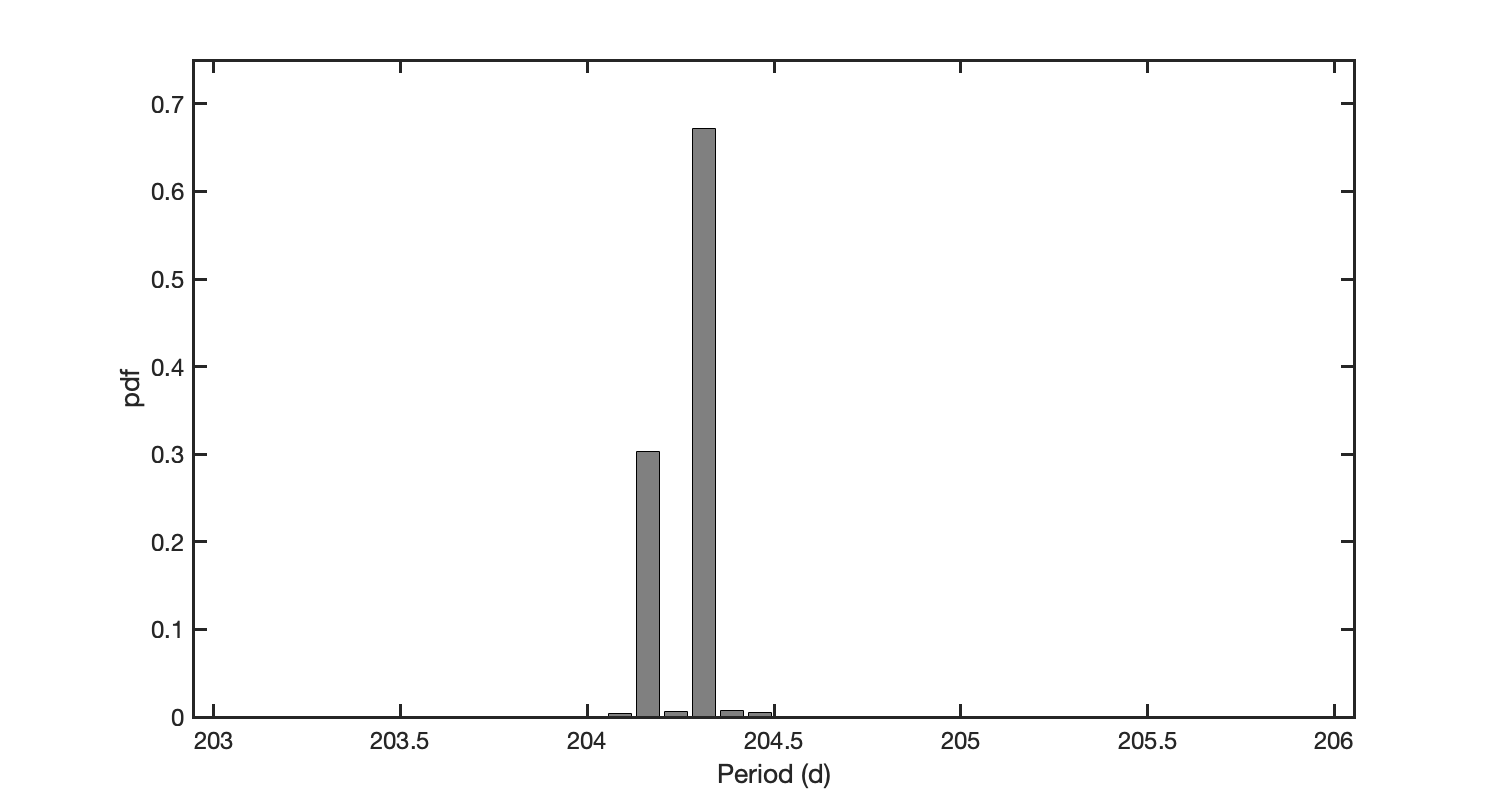}
    \caption{Marginal posterior of the period obtained from the AIS algorithm using a uniform prior.}
    \label{fig:marginalPosteriorUniform}
\end{figure}

We analyzed the frequencies in the time series of GJ~3512. Figure~\ref{fig:periodograms} shows the periodogram of the radial velocity curve in Fig.~\ref{fig:FBS_real}. The figure compares the Lomb-Scargle periodogram (top panel) and the l1-periodogram described in \citet{Hara2017}. Both identify the main period of the time series at $\sim 203.5$~d and a number of harmonics. For comparison with the methodology presented in this article, we chose a uniform prior in the range [195,210] (selected from the pdf in Fig.~\ref{fig:posteriorP_real}). We then approximated the marginal posterior distribution of the period in the model using an AIS algorithm, similar to Sect.~\ref{sec:realdata}. The result is shown in Fig.~\ref{fig:marginalPosteriorUniform}.  {The distribution has a mean $\mu = 204.37$~d and a standard deviation $\sigma = 0.13$. The figure shows the overall distribution, which does not reveal the harmonics, but only the fundamental period.} 

 {The distribution obtained with the uniform prior seems to contain less information than the distribution shown in the right panel of Fig.~\ref{fig:posteriorP_real}. More precisely, it does not accurately approximate the marginal posterior distribution for $P$.
This is due to the low a priori information provided by the uniform prior when used in Bayesian inversion problems.}

\section{Conclusions}
\label{sec:conclusions}

In this paper, we show an approach to Bayesian inference in parametric models with highly concentrated posterior distributions. In particular, we focus on oscillatory or quasi-oscillatory models where the parameter associated with the period (either frequency or period) of the oscillation governs the {joint} posterior distribution. In these models, it is necessary to construct informative prior distributions to reduce the parameter space to be explored. More specifically, it is necessary that the prior distribution of the parameter associated with the period of the oscillation be informative, so that the inference algorithm correctly explores the parameter space. Historically, a frequency analysis has been performed on the data or observations to reveal the frequencies of the oscillations. This information is then used to construct the prior distribution of the parameter associated with the period. 

We propose to {approximate the posterior distribution} for the hyperparameters of a periodic kernel of a {GP} fitted to the observations. Subsequently, the posterior distribution of the hyperparameter period is used as the prior distribution of the oscillation period in the parametric model. While this prior distribution is constructed from the data that is then used for the inference, it is sufficiently diffuse to make the inference unbiased. Note that only the prior of the oscillation period is constructed in this way, while those of the remaining model parameters are constructed in an independent way.

 {Examples of astrophysical applications include radial velocity curves of spectroscopic binaries or exoplanet host stars, light curves of spectroscopic and eclipsing binaries, pulsating stars, stars with oscillations triggered by flaring events, and gravitational wave data.}
In Sect.~\ref{sec:applications}, we first show with simulations what the posterior distribution of the hyperparameters of a basic periodic kernel obtained from the full Bayesian solution looks like and how this differs from a GP regression with optimization. Subsequently, we show the result of applying our method to real data and how it considerably improves the inference results with respect to using an uninformative prior for the period of the oscillation.  {We applied the method to the radial velocity curve of the star GJ~3512 and to the optical light curve of the star HD~112661. The results show that the combination of the periodogram with a GP/AIS method, as the one proposed here, performs better in terms of approximating the actual posterior distribution. To demonstrate this point, we compare the results of the methodology proposed by us with those of using prior distributions constructed from the frequency analysis of the data.} 

Therefore, the construction of data-driven informative priors for the period in periodic or quasi-periodic models seems the best approach to guarantee a large concentration rate of the posterior distribution {approximated with computational tools}.  {In turn, this approach {avoids} errors in estimating the model parameters due to the poor approximation of the marginal posterior distribution of the period.}

\begin{acknowledgments}
This work was supported by grants PID2019-105032GB-I00, PID2020-116683GB-C21, PID2024-158181NB-I00 and PID2024-159557OB-C21, funded by MCIN/AEI/10.13039/501100011033 and by ``ERDF A way of making Europe'', and Community of Madrid (project TEC-2024/COM-89 IDEA-CM). J.L-S. acknowledges partial support from the Office of Naval Research (award no. N62909-24-1-2095). The work was also partially supported by the Young Researchers R$\&$D Project, ref. num. F861 (AUTO-BA-GRAPH) funded by Community of Madrid and Rey Juan Carlos University.
\end{acknowledgments}

\begin{contribution}

JLS led the analysis of both simulations and applied case studies, developed the Gaussian process regression codebase, and drafted and submitted the manuscript. LM devised a rigorous mathematical framework and executed the optimization of the underlying algorithms. JM validated the proposed mathematical formalism and independently verified the application results. GVV implemented the frequency‐analysis software and confirmed the accuracy of its outputs.


\end{contribution}

%



\bibliographystyle{aasjournalv7}
\bibliography{GP}



\end{document}